\documentclass[3p,times,procedia]{elsarticle}
\usepackage{ecrc}
\usepackage[bookmarks=false]{hyperref}
    \hypersetup{colorlinks,
      linkcolor=blue,
      citecolor=blue,
      urlcolor=blue}

\volume{00}

\firstpage{1}

\journalname{Procedia Computer Science}

\runauth{Salgado et al.}

\jid{dummy}       
\jnltitlelogo{}

\usepackage{amssymb}
\usepackage{float}
\usepackage[utf8]{inputenc}

\usepackage[figuresright]{rotating}

\begin{document}
\begin{frontmatter}


\dochead{The 3rd International Workshop on Causality, Agents and Large Models (CALM-26)\\ April 14-16, 2026, Istanbul, Türkiye}

\title{Causal Discovery for Explainable AI: A Dual-Encoding Approach}

\author[utep]{Henry Salgado\corref{cor1}}
\author[utep2]{Meagan R. Kendall}
\author[utep]{Martine Ceberio}

\address[utep]{Department of Computer Science, The University of Texas at El Paso, El Paso, TX 79968, USA}
\address[utep2]{Department of Engineering Education and Leadership, The University of Texas at El Paso, El Paso, TX 79968, USA}

\begin{abstract}
Understanding causal relationships among features is fundamental for explaining machine learning model decisions. However, traditional causal discovery methods face challenges with categorical variables due to numerical instability in conditional independence testing. We propose a dual-encoding causal discovery approach that addresses these limitations by running constraint-based algorithms with complementary encoding strategies and merging results through majority voting. Applied to the Titanic dataset, our method identifies causal structures that align with established explainable methods. 
\end{abstract}


\cortext[cor1]{Corresponding author. Tel.: +1-915-892-0496; fax: N/A.}

\end{frontmatter}

\email{hsalgado@miners.utep.edu}



\section{Introduction}

Machine learning models are increasingly being used in critical decision-making contexts, from medical diagnosis to criminal justice \cite{verma_fairness_2018}. Although these models achieve high predictive accuracy, their opacity raises concerns about trustworthiness and accountability \cite{joshi2025fairness}. Existing explainability methods such as SHAP \cite{lundberg_unified_2017} and LIME \cite{ribeiro_why_2016} identify which features influence predictions but fail to explain \emph{why} certain features matter or \emph{how} they interact causally. Causal discovery methods offer a promising alternative by inferring directional relationships from observational data \cite{glymour_review_2019, bystrova_causal_2024}. However, even between variables, there can be important interactions. For example, age might influence likelihood of disease, but we might wonder specifically whether young or older people are more affected. 

This need for granular causal understanding is compounded by the fact that real-world datasets are complex with mixed data types that contain continuous and categorical variables. Standard approaches like one-hot encoding can lead to singular covariance matrices that cause conditional independence tests to fail. We address these challenges through a dual-encoding causal discovery approach by executing constraint-based algorithms with two complementary encoding strategies and merging results via majority voting. Our experiments on the Titanic dataset demonstrate that this approach produces unified causal graphs that function as global explanation mechanism.

To situate our contribution, we first review existing approaches to model explanation (\S\ref{sec:explanation}) and their limitations in capturing causal relationships. We then examine causal discovery methods (\S\ref{sec:causal_discovery}) that could address these limitations, before identifying the specific technical challenges posed by mixed data types that motivate our dual-encoding strategy (\S\ref{sec:method}).

\section{Background}

\subsection{Explanation Methods}
\label{sec:explanation}

Given the limitations of black-box models discussed above, various methods have been proposed to explain model predictions. Global explanation methods aim to characterize overall model behavior by aggregating feature contributions across all instances. Methods like SHAP \cite{lundberg_unified_2017} compute average feature importance, while Partial Dependence Plots \cite{altwaijry_improved_2026} reveal marginal effects of individual features. However, these approaches have fundamental limitations: (1) they capture \emph{associations} rather than causal relationships, making it unclear whether a feature directly influences the outcome or merely correlates with true causes; (2) they treat features as independent units, missing important interactions and mediating relationships; and (3) they provide no information about directionality, whether feature A causes B or vice versa. These limitations of association-based explanation methods motivate the use of causal discovery, which explicitly models directional relationships between variables.

\subsection{Causal Discovery Methods}
\label{sec:causal_discovery}

Causal discovery algorithms infer directional relationships from observational data. Constraint-based methods, such as the PC algorithm and Fast Causal Inference (FCI) \cite{glymour_review_2019}, rely on conditional independence testing to identify causal structure. These algorithms assume the Causal Markov Condition (each variable is independent of its non-descendants given parents), Faithfulness (observed independencies match graph structure), and often Causal Sufficiency (no hidden confounders). Constraint-based discovery algorithms generally proceed in two phases: first identifying adjacencies via conditional independence tests, then orienting edges using structural rules. A critical orientation pattern is the \emph{unshielded collider} (or v-structure): \( X \rightarrow Y \leftarrow Z \), where two non-adjacent variables \( X \) and \( Z \) both influence a common effect \( Y \). Such structures are identifiable because \( X \) and \( Z \) are marginally independent but become dependent when conditioning on \( Y \). Once colliders are oriented, propagation rules orient remaining edges while avoiding new colliders or cycles. PC assumes causal sufficiency, whereas FCI relaxes this assumption by introducing bidirected edges (\( X \leftrightarrow Y \)) to represent the possible uncertainty of latent confounders, providing a more conservative representation when hidden variables may be present. 

However, applying these causal discovery methods to real-world datasets with mixed data types presents significant challenges. The encoding choices required for conditional independence testing create a trade-off: preserving categorical distinctions (for value-specific insights) versus ensuring numerical stability (for valid statistical tests). This tension motivates our dual-encoding approach.

\section{Methods}
\label{sec:method}

\subsection{Dual-Encoding Strategy}

We now formalize our approach to the mixed data type challenge. Conditional independence testing in constraint-based algorithms typically employs Fisher's \(z\)-test, which assumes continuous variables and estimates partial correlations from the covariance matrix. When categorical variables are one-hot encoded, the resulting design matrix becomes rank-deficient (the sum of dummy variables equals a constant), producing a singular covariance matrix that cannot be inverted for the \(z\)-test. To resolve this issue, we propose running causal discovery twice with complementary encoding strategies:

\begin{enumerate}
\item \textbf{Drop-first encoding:} Remove the first category from each categorical variable
\item \textbf{Drop-last encoding:} Remove the last category from each categorical variable
\end{enumerate}

Each encoding yields a different invertible covariance matrix, potentially revealing different conditional independencies. By running FCI on both encodings and merging results, we capture edges that are consistent across encoding choices.

\subsection{Data Preparation}

Continuous features are discretized using supervised entropy-based binning to identify value ranges that maximize information gain relative to the outcome. This transformation enables causal discovery algorithms to detect nonlinear relationships while maintaining interpretability. For example, \textit{Age} might be discretized into intervals like [0,5), [5,30), [30,inf) representing children, adults, and elderly passengers.

\subsection{Graph Merging and Correlation Weighting}

Our method proceeds in three main stages:

\textbf{Step 1: Causal Discovery.} We incorporate domain knowledge by enforcing the outcome variable as a sink node, ensuring that all directed paths terminate at the target. This constraint reflects the assumption that, in observational settings, outcome variables do not causally influence feature values. We then apply the Fast Causal Inference (FCI) algorithm using the \texttt{causal-learn} \cite{zheng2024causal} library with significance level \(\alpha=0.01\) and Fisher's \(z\)-test to each encoded dataset (drop-first and drop-last). 

\textbf{Step 2: Graph Merging.} The two encoding-specific causal graphs are merged via majority voting. An edge is retained in the unified graph if it appears in at least one individual graph. When an edge appears in both graphs, we require consistent orientation (e.g., if graph 1 has \(A \rightarrow B\) and graph 2 has \(A \rightarrow B\), the unified graph includes \(A \rightarrow B\); conflicting orientations are resolved by retaining the undirected edge).

\textbf{Step 3: Correlation Weighting.} For each retained edge, we compute Pearson correlation coefficients to quantify association strength and direction. Edges are labeled as \emph{supportive} (positive correlation, \(r > 0\)) or \emph{opposing} (negative correlation, \(r < 0\)). To capture mutual exclusivity among discretized ranges derived from the same continuous feature (e.g., Age[0,5) vs. Age[30,inf)), we assign a fixed negative correlation of \(r = -1\) between such subsets, reflecting their inherent incompatibility.

\section{Experiments and Results}

\subsection{Datasets}

We evaluated our approach on a popular benchmark classification dataset. \textbf{Titanic Dataset:} 891 passengers with features including Pclass (categorical ticket class), Sex (binary), Age (continuous), SibSp (count of siblings/spouses aboard), Parch (count of parents/children aboard), Fare (continuous), and Embarked (categorical port of embarkation), predicting survival (binary). We selected this dataset because it contains mixed data types—combining categorical, binary, and continuous variables—which allows us to demonstrate our method's handling of heterogeneous features.

\subsection{Titanic Results}

\begin{figure}[H]
\centering
\includegraphics[width=0.35\columnwidth]{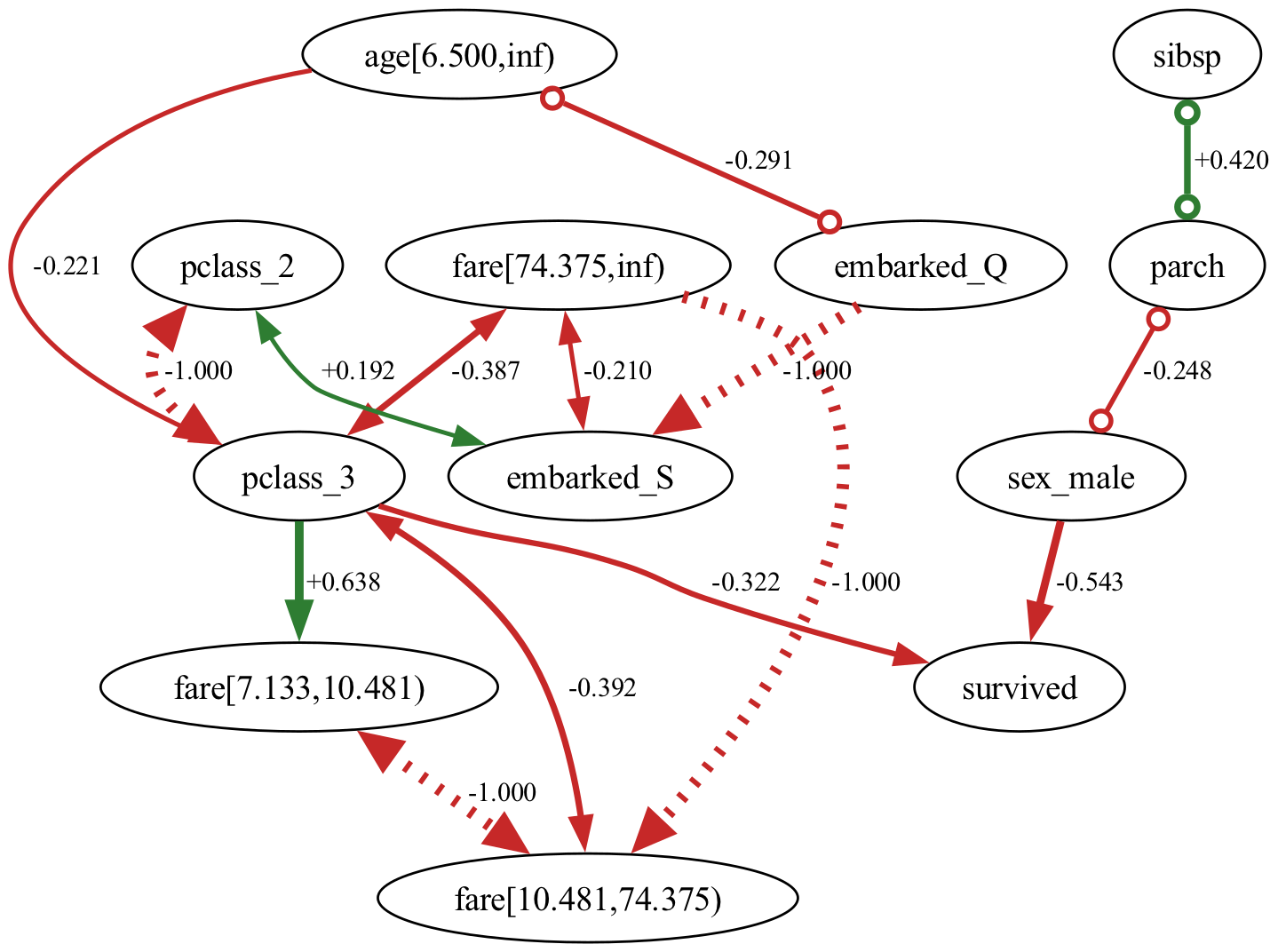}
\includegraphics[width=0.35\columnwidth]{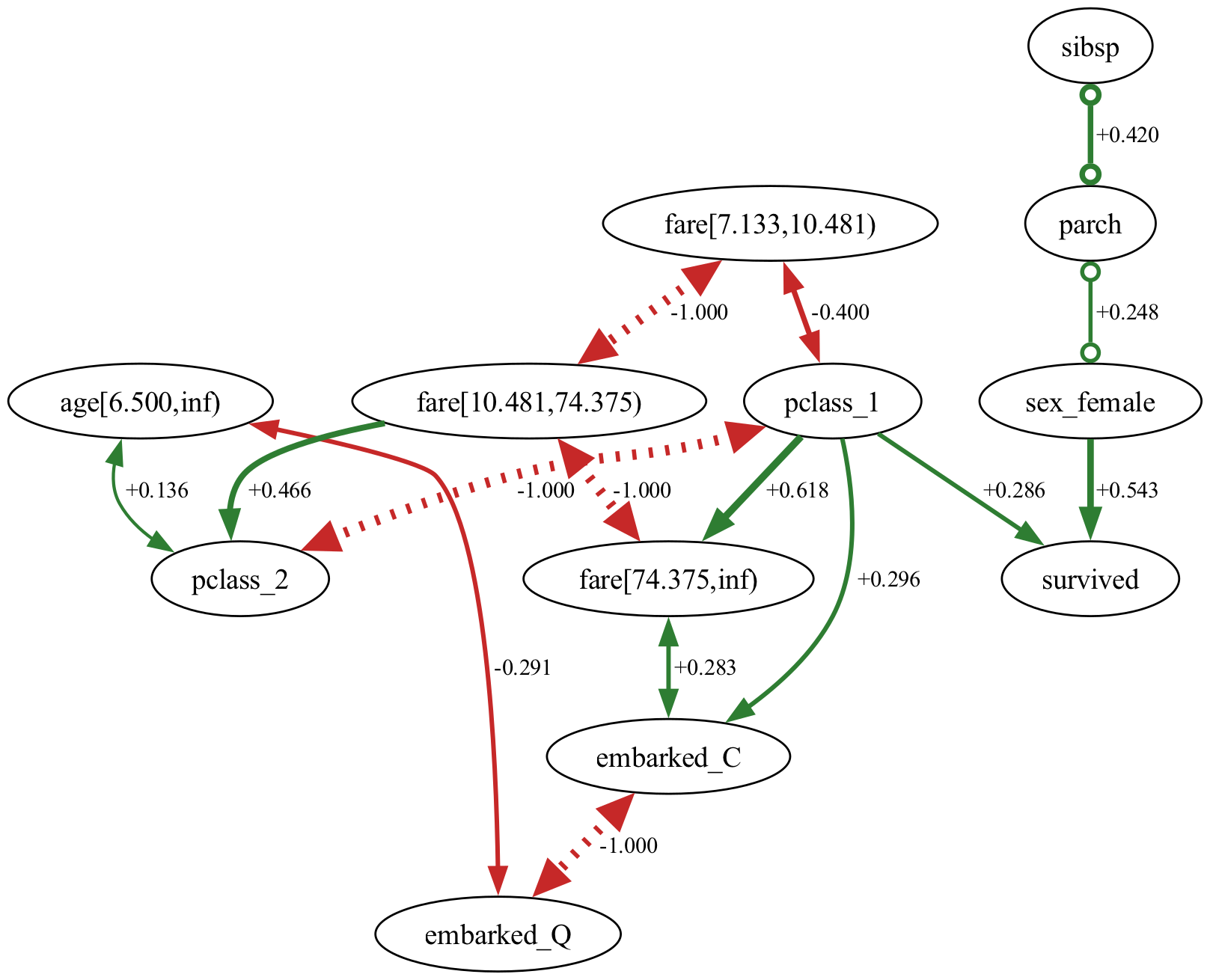}
\caption{Causal graphs for the Titanic dataset using drop-first (left) and drop-last (right) encoding strategies. Structural consistency across encodings demonstrates robustness to categorical reference choice.}
\label{fig:titanic_graphs}
\end{figure}

Figure~\ref{fig:titanic_graphs} shows the causal graphs obtained under the two encoding strategies. 
Despite different reference categories, both encodings consistently identify key relationships: 
\textit{Sex} directly influences \textit{Survived} (strong negative correlation for males), higher 
\textit{Pclass} positively influences survival, and \textit{Fare} mediates the relationship between 
class and survival.

\begin{figure}[H]
\centering
\includegraphics[width=0.35\columnwidth]{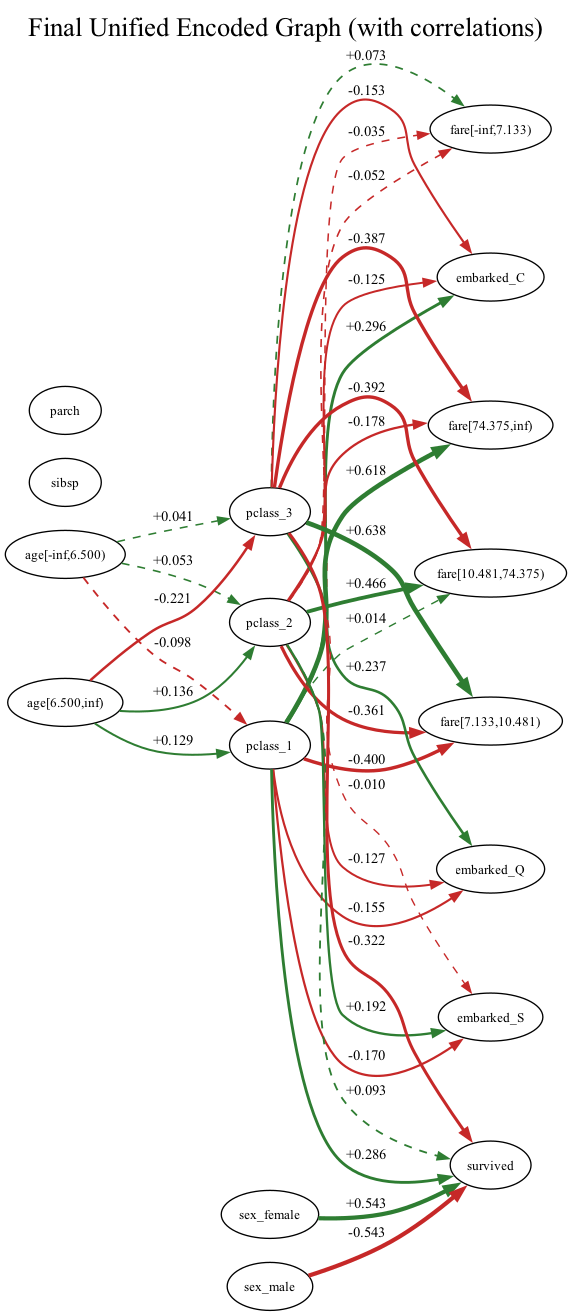}
\caption{Unified causal graph for the Titanic dataset obtained by merging encoding-specific graphs via majority voting and correlation weighting.}
\label{fig:titanic_unified}
\end{figure}

The unified graph produced by merging both encodings (Fig.~\ref{fig:titanic_unified}) contains 
17 feature nodes, with 15 positively correlated and 17 negatively correlated edges. The resulting structure highlights \textit{Sex}, \textit{Pclass}, and \textit{Age} as central causal drivers of survival outcomes.

\subsection{Validation Against Baseline Methods}

To assess whether the discovered causal relationships correspond to established notions of feature 
importance, we compared our results against SHAP explanations and pruned decision trees. Both models achieved good predictive performance on the Titanic dataset, with the decision tree reaching 82\% accuracy and the random forest model (used for SHAP analysis) achieving 84\% accuracy.

For the Titanic dataset, the decision tree splits first on \textit{Sex}, followed by 
\textit{Pclass} and \textit{Age}, mirroring the dominant causal pathways identified in the unified 
graph (Fig.~\ref{fig:titanic_unified}). This structural agreement indicates that the causal discovery process captures genuinely discriminative relationships. 
\begin{figure}[H]
\centering
\includegraphics[width=0.50\columnwidth]{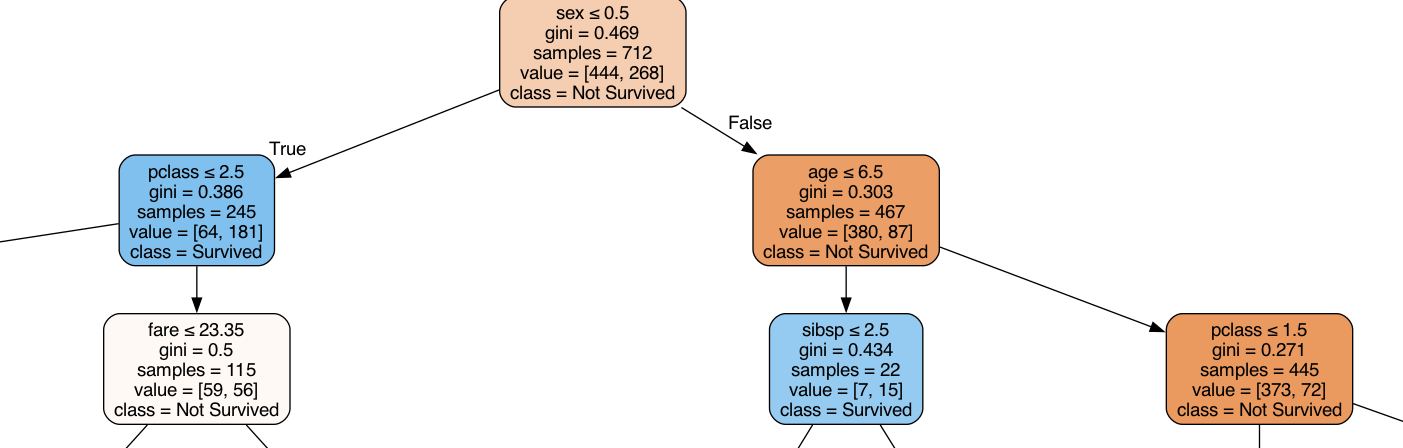}
\caption{Pruned decision tree trained on the Titanic dataset. The root split on \textit{Sex}, followed by \textit{Pclass} and \textit{Age}, aligns with the causal relationships identified by FCI.}
\label{fig:titanic_decision}
\end{figure}

\begin{figure}[H]
\centering
\includegraphics[width=0.40\columnwidth]{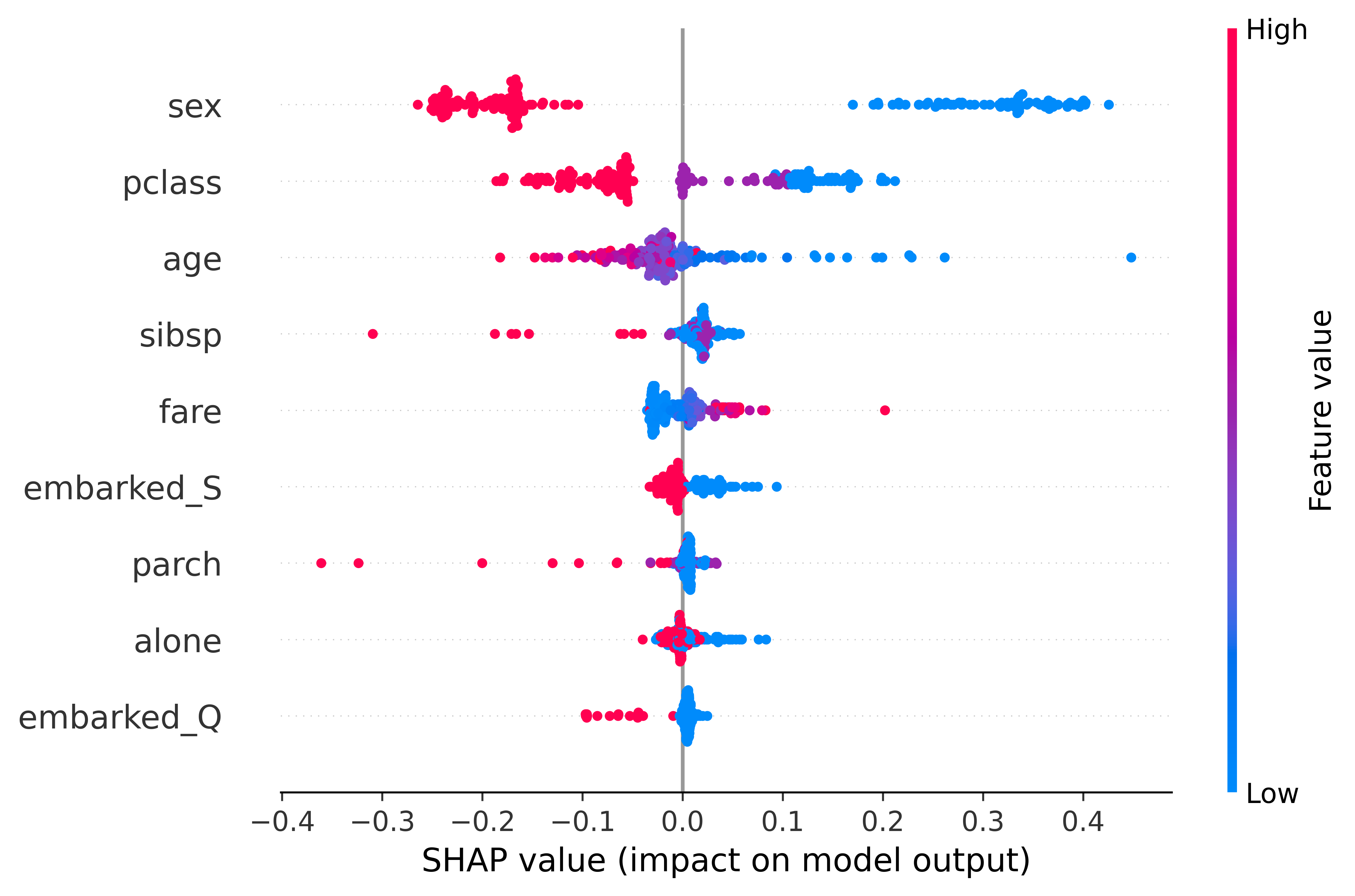}
\caption{SHAP summary plot illustrating global feature importance for the Titanic dataset. The most influential features align with those identified in the unified causal graph.}
\label{fig:titanic_shap}
\end{figure}

Figure~\ref{fig:titanic_shap} further confirms this alignment through SHAP-based global feature 
importance, where \textit{Sex}, \textit{Pclass}, and \textit{Age} emerge as the most influential 
features.

\section{Discussion}

This work advances the idea of subpopulation-level causal interpretability by providing a pragmatic robustness layer on top of classical PC/FCI workflows. By running FCI with complementary encodings and merging results, we obtain causal graphs that are stable across encoding choices and aligned with domain knowledge.The discovered causal structures provide several advantages over correlation-based feature importance: (1) directionality reveals which features influence others versus shared causes, (2) mediating relationships identify indirect effects, and (3) causal graphs support counterfactual reasoning about interventions.
Future work will integrate these causal structures with argumentation frameworks to generate instance-level explanations that capture both global causal relationships and local feature interactions.

\subsection{Limitations}

Our approach inherits limitations of constraint-based causal discovery: assumptions of Causal Markov, Faithfulness, and (for FCI) relaxed sufficiency must hold approximately. Discretization introduces binning choices that affect discovered structure. The method also requires sufficient sample size for reliable conditional independence testing. Additionally, we have not conducted controlled comparisons against modern mixed-data causal discovery methods, quantitative stability assessments or validation on synthetic data with known ground truth.

\section{Conclusion and Future Work}

We presented a dual-encoding causal discovery approach that addresses numerical instability when applying constraint-based algorithms to datasets with categorical variables. Validation on Titanic dataset demonstrates that merged causal graphs align with domain knowledge and established feature importance methods. Future work will explore alternative causal discovery algorithms beyond FCI, investigate the sensitivity of discovered structures to different significance thresholds (p-values), and integrate these causal graphs with argumentation frameworks to generate instance-level explanations that leverage both global causal relationships and local feature interactions.

\bibliographystyle{elsarticle-num}
\bibliography{references}

\end{document}